\documentclass{scrartcl}
\usepackage[utf8]{inputenc}

\title{Macro F1 and Macro F1}
\subtitle{a note}
\author{Juri Opitz \\ \and Sebastian Burst}
\date{}

\usepackage{graphicx}
\usepackage{amsmath}
\usepackage{amssymb}
\usepackage{amsthm}
\usepackage{comment}
\usepackage[bf]{caption}
\usepackage{bm}
\usepackage{booktabs}
\usepackage{wrapfig}
\usepackage[toc,page]{appendix}
\usepackage[usenames,dvipsnames]{xcolor}
\usepackage{fancyvrb}
\usepackage{mdframed}
\usepackage{url}
\usepackage{pgf,tikz}
\usepackage{adjustbox}
\usepackage{stmaryrd}
\newmdtheoremenv{theo}{Theorem}
\definecolor{calpolypomonagreen}{rgb}{0.12, 0.3, 0.17}
\usepackage{mathtools}
\usepackage{hyperref}
\usepackage{subcaption}
\begin{document}

\maketitle


\begin{abstract}
    The `macro F1' metric is frequently used to evaluate binary, multi-class and multi-label
    classification problems. Yet, we find that there exist two different formulas to calculate this quantity. In this note, we show that only under rare circumstances the two computations can be considered equivalent. More specifically, one formula  well `rewards' classifiers which produce a skewed error type distribution. In fact, the difference in outcome of the two computations can be as high as 0.5. The two computations may not only diverge in their scalar result but can also lead to different classifier rankings.
\end{abstract}

\section{Introduction}
We find two formulas which are used to compute `macro F1'. We name them `averaged F1' and `F1 of averages'.

\paragraph{Preliminaries} For any classifier  $f : D \rightarrow C=\{1,...,n\}$ and finite set $S \subseteq D\times C $, let  $m^{f,S}\in\mathbb{N}_0^{n\times n}$ be a confusion matrix, where $m^{f,S}_{ij}=|\{s\in S~|~f(s_1)=i \land  s_2=j\}|$. We omit superscripts whenever possible.
For any such matrix, let $P_i, R_i$ and $F1_i$ denote precision, recall and F1-score with respect to class $i$:
\begin{equation}
P_i=\frac{m_{ii}}{\sum_{x=1}^n m_{ix}} \text{ ; }
R_i = \frac{m_{ii}}{\sum_{x=1}^n m_{xi}} \text{ ; }
F1_i = H(P_i, R_i) = \frac{2P_iR_i}{P_i+R_i}
\end{equation}
with $P_i,R_i,F1_i=0$ when the denominator is zero. $H$ is the harmonic mean. Precision and recall are also known as positive predictive value and sensitivity.

\paragraph{Averaged F1: arithmetic mean over harmonic means} F1 scores are computed for each class and then averaged via arithmetic mean:\footnote{Three among many examples: \cite{wu2017unified,lipton2014optimal,rosenthal2015semeval}}
\begin{equation}
    \label{eq:avgf1}
    \mathcal{F}_1 = \frac{1}{ n }\sum_x\text{F}1_x = \frac{1}{ n }\sum_x \frac{2P_x  R_x}{P_x+R_x}.
\end{equation}

\paragraph{F1 of averages: harmonic mean over arithmetic means} The harmonic mean is computed over the arithmetic means of precision and recall:\footnote{Some among many examples: \cite{sokolova2009systematic, rudinger-etal-2018-neural,santos2011comparative,opitz-frank-2019-argument} and also \url{http://rushdishams.blogspot.com/2011/08/micro-and-macro-average-of-precision.html}: \textit{``take the average of the precision and recall (...) The Macro-average F-Score will be simply the harmonic mean of these two figures.''}}
\begin{equation}
\label{eq:f1ofavg}
    \mathbb{F}_1 = H(\bar{P},\bar{R})
    =\frac{2\bar{P}\bar{R}}{\bar{P}+\bar{R}}
    =2\frac{(\frac{1}{ n }\sum_x  P_x) (\frac{1}{ n } \sum_x R_x)}{\frac{1}{ n }\sum_x P_x+\frac{1}{ n }\sum_x R_x}
\end{equation}
We already see an important difference between these two definitions: In $\mathbb{F}_1$, the precision values of each class are multiplied with the recall values of all other classes. In $\mathcal{F}_1$, the precision of each class is multiplied only with the recall of the same class.

In the remainder of this paper, we first present a mathematical analysis of the two formulas and then consider some practical implications.

\section{Mathematical analysis}
\paragraph{Theorem} $\forall m\in \mathbb{N}_0^{n\times n}$:
\begin{enumerate}
	\item $\mathbb{F}_1 \geq  \mathcal{F}_1$ 
	\item $\mathbb{F}_1 > \mathcal{F}_1\Leftrightarrow $
	$	\exists i \in C: P_i\neq R_i \Leftrightarrow
		\exists i,j \in C: P_i < R_i, P_j > R_j$ 
	\item $$
		\sup_{m\in \mathbb{N}_0^{n\times n}}
		\{\mathbb{F}_1-\mathcal{F}_1\} =
		\begin{cases}
			0.5,n\text{ is even}\\
    		0.5 - \frac{1}{2n^2},\text{else}
    	\end{cases}$$ 
\end{enumerate}
The first property follows directly from the next Lemma. Proofs for (2.), (3.) and the following Lemma are in the appendix.

\paragraph{Lemma} $\forall m\in \mathbb{N}_0^{n\times n}$ not a hollow matrix:
\begin{equation}
\label{eq:diff}
     \Delta =  \mathbb{F}_1 - \mathcal{F}_1 =  \frac{1}{ n \sum_{x \in C}(P_x +R_x)} \sum\limits_{\substack{x,y \in C\\ P_x+R_x, P_y+R_y \neq 0}} \frac{(P_x R_y - P_y R_x)^2 }{(P_x +R_x)(P_y +R_y)}
\end{equation}

\noindent Less formally, \begin{itemize}
    \item $\Delta$ is large when there are many classes with $|P-R|\gg0$. However, $\Delta$ does not necessarily increase monotonously when $|P-R|$ is increased for single classes, because all possible class pairs need to be considered.
    \item $\Delta$ is maximised when there are classes with $(P,R) \rightarrow (1,0)$ and other classes with $(P,R) \rightarrow (0,1)$.
    Then, for all classes F$1 \rightarrow 0$ $(\Rightarrow \mathcal{F}_1\rightarrow0)$ and $\bar{P},\bar{R}\approx0.5$ $(\Rightarrow\mathbb{F}_1\approx0.5)$.
\end{itemize}
We can summarise that a large difference in outcomes is encountered in situations where a classifier has a strong bias towards certain types of errors (e.g., in the binary case, frequent/infrequent type I/II errors)
because in such cases, not all classes will share the same bias (Theorem, 2.).
$\mathbb{F}_1$ `rewards' such classifiers.
Note that while different error type distributions might be desirable in certain applications (e.g. high recall for some classes and high precision for other classes), $\mathbb{F}_1$ is insensitive to which classes have which distribution.

\section{Numerical experiments}
Before we analyse what $\Delta$ can be expected in average cases, we want to highlight that the two metrics may not only differ in their absolute value but can also yield different classifier rankings. That is, when a classifier outperforms another classifier on a fixed data set according to one metric, it may at the same time be worse w.r.t. the other metric. Consider Tables \ref{tab:subtab1} and \ref{tab:subtab2}: Introducing a bias towards class b improves $\mathbb{F}_1$, impairs $\mathcal{F}_1$.
\begin{table}[!htb]
\begin{minipage}{.5\linewidth}
    \centering
    \begin{tabular}{c|c|c}
            & a & b \\
        \hline
        a   & 5 & 10 \\
        b   & 5 & 10 \\
    \end{tabular}
    \caption{$\mathbb{F}_1$ = 0.5, $\mathcal{F}_1$ = 0.49}
    \label{tab:subtab1}
\end{minipage}%
\begin{minipage}{.5\linewidth}
    \centering
    \begin{tabular}{c|c|c}
            & a & b \\
        \hline
        a   & 1 & 1 \\
        b   & 9 & 19 \\
    \end{tabular}
    \caption{$\mathbb{F}_1$ = 0.55, $\mathcal{F}_1$ = 0.48}
    \label{tab:subtab2}
\end{minipage}
\label{tab:ex1}
\end{table}

\begin{figure}
\centering
\begin{subfigure}{.5\textwidth}
  \centering
  \includegraphics[width=1.0\linewidth]{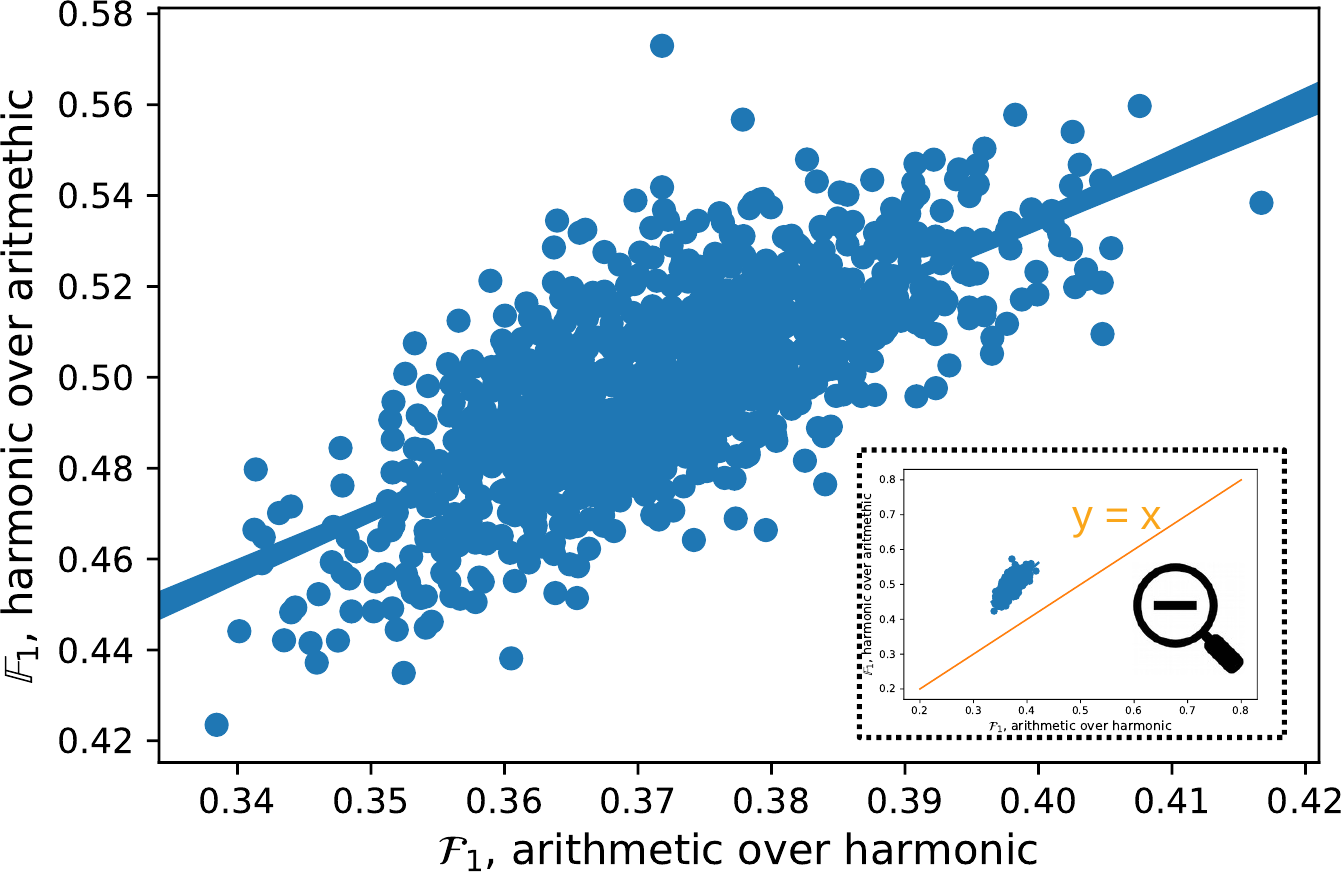}
  \caption{}
  \label{fig:test1}
\end{subfigure}%
\begin{subfigure}{.5\textwidth}
  \centering
  \includegraphics[width=1.1\linewidth]{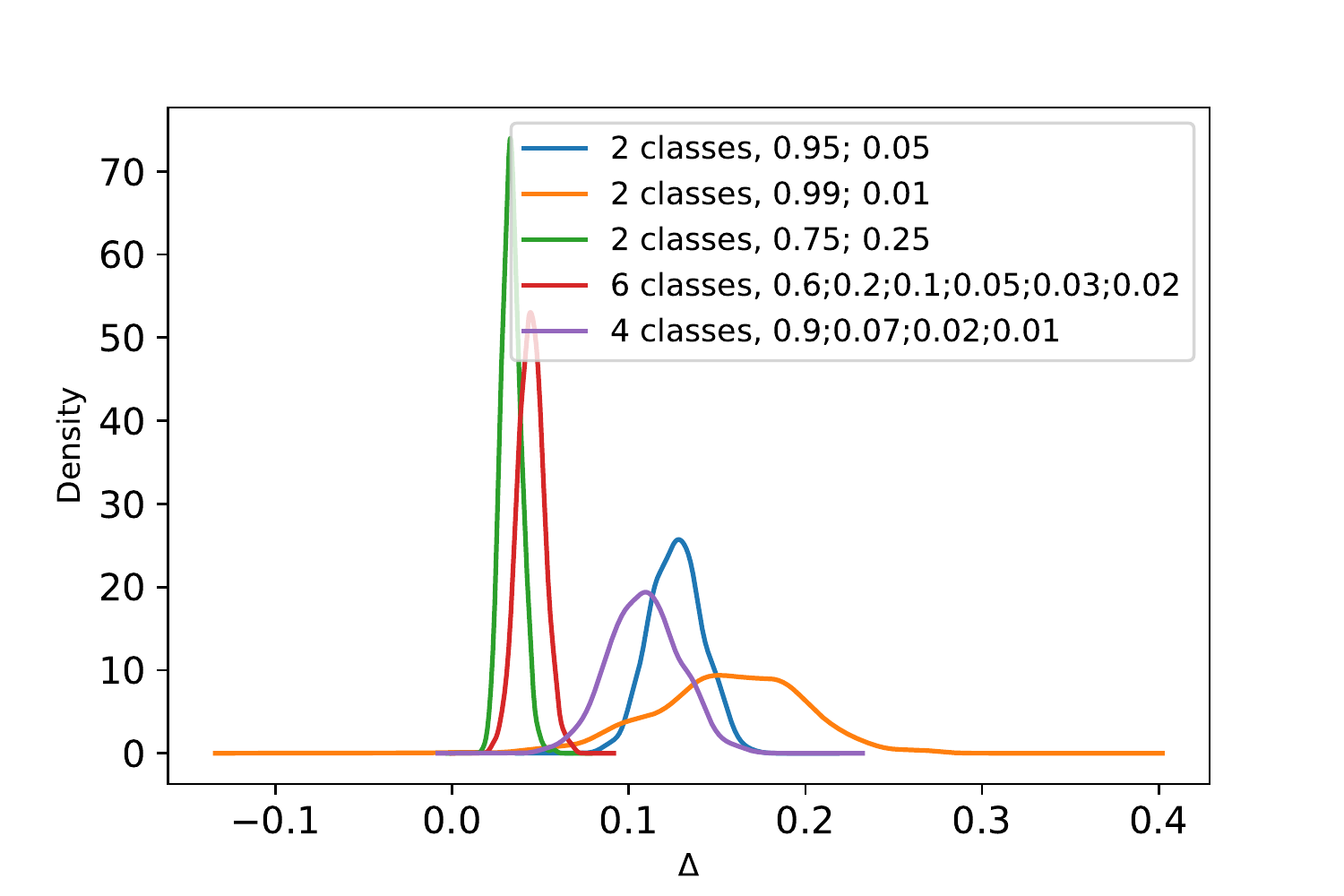}
  \caption{}
  \label{fig:test2}
\end{subfigure}
\caption{\textbf{(a)}: Macro F1 results from 1000 randomly sampled binary classification tasks with class distribution 95\% vs.\ 5\% and random classifier. \textbf{(b)}: $\Delta$ density (KDE) from 1000 randomly sampled classification tasks with respect to various class distributions (see legend) and random classifier.}
\end{figure}

\paragraph{Bias in data} `Macro F1' is often used in situations where classes are unevenly distributed. Figures \ref{fig:test1} (binary) and \ref{fig:test2} (multi-class) show classifier results on 1,000 random data sets S with 1,000 data examples each, where the `true' label is drawn from a multinomial probability distribution (see legend). We solve these tasks with `dummy'-classifiers $f$ that predict classes uniformly at random.\footnote{Such classifiers are frequently chosen as a baseline by researchers.}
Consider the binary classification results in Figure \ref{fig:test1}: First, the harmonic mean over arithmetic means ($\mathbb{F}_1$) indeed is more benevolent towards the classifiers (maximum appr.\ 0.56) while the arithmetic mean over harmonic means ($\mathcal{F}_1$) yields more conservative results (maximum score appr.\ 0.41). The root mean squared deviation $\sqrt{1000^{-1}\sum_{(f,S)}(\mathbb{F}_1(m^{f,S}) - \mathcal{F}_1(m^{f,S}))^2}$ is 0.13. Second, while there appears to be a solid correlation between the two macro F1 metrics, it is by no means perfect (Pearson's $\rho = 0.72, p<0.0001$; Spearman's $\rho = 0.69, p<0.0001$) and allows for different classifier rankings.

\begin{figure}
    \begin{subfigure}{.5\linewidth}
      \centering
      \includegraphics[width=\textwidth]{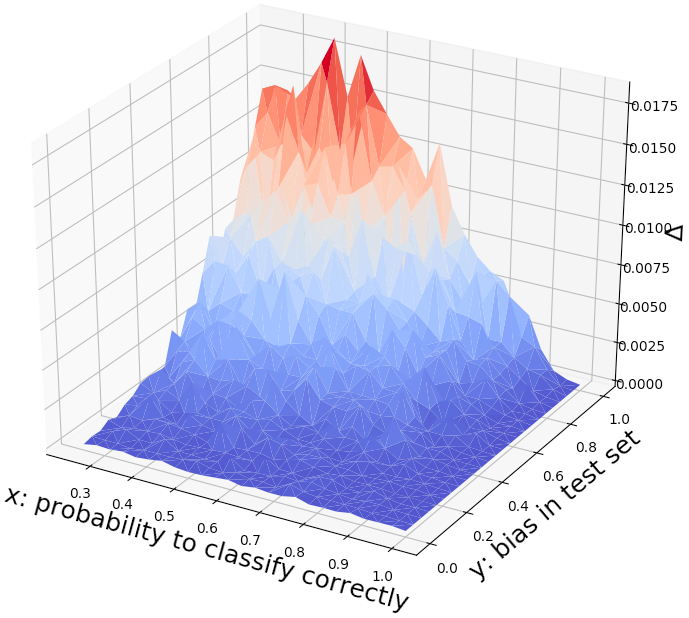}
    \caption{n=4}
    \label{fig:sf1}
    \end{subfigure}%
    \begin{subfigure}{.5\linewidth}
      \centering
      \includegraphics[width=\textwidth]{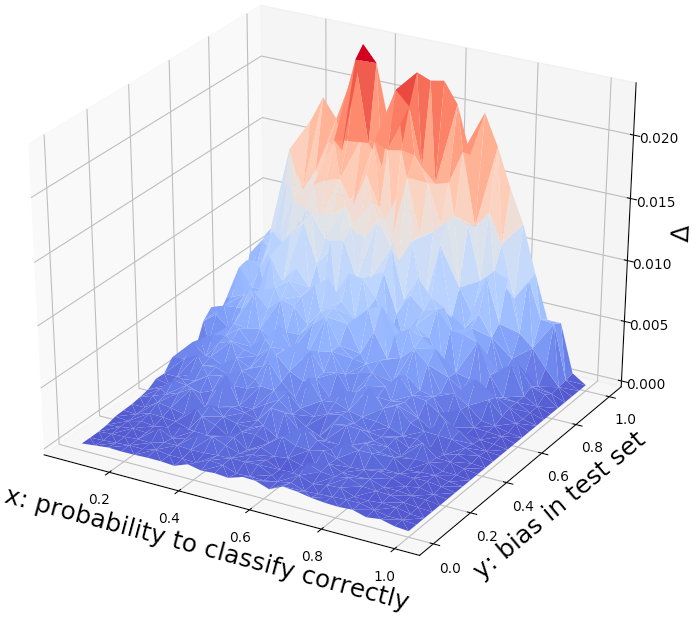}
    \caption{n=13}
    \label{fig:sf2}
    \end{subfigure} 
    \caption{ $\Delta$ for random classification tasks on 2000 data points with different classifier performance ($x$-axis) and different label distribution ($y$-axis).}
     \label{fig:ranclas}
\end{figure}

Figures \ref{fig:sf1} and \ref{fig:sf2} show $\Delta$ for random classification tasks with varying classifier performance and label distribution.
The $x$-axis represents the probability that data points are classified correctly (ranging from $\frac{1}{n}$ to 1, with the  remaining probability evenly distributed over remaining classes).
With the $y$-axis we control the class distribution in the data set (the proportion of data points for class $i$ ranges from $\frac{1}{n} [y=0]$ to $i\frac{1}{\sum_i i}[y=1]$).
Note that this is a much weaker bias than before.
While both $\mathbb{F}_1$ and $\mathcal{F}_1$ are roughly proportional to $x$, we still find differences up to 2 percentage points whenever the classifier's accuracy is not 1 and the data set is skewed.

\begin{figure}
    \begin{subfigure}{.5\linewidth}
      \centering
      \includegraphics[width=\textwidth]{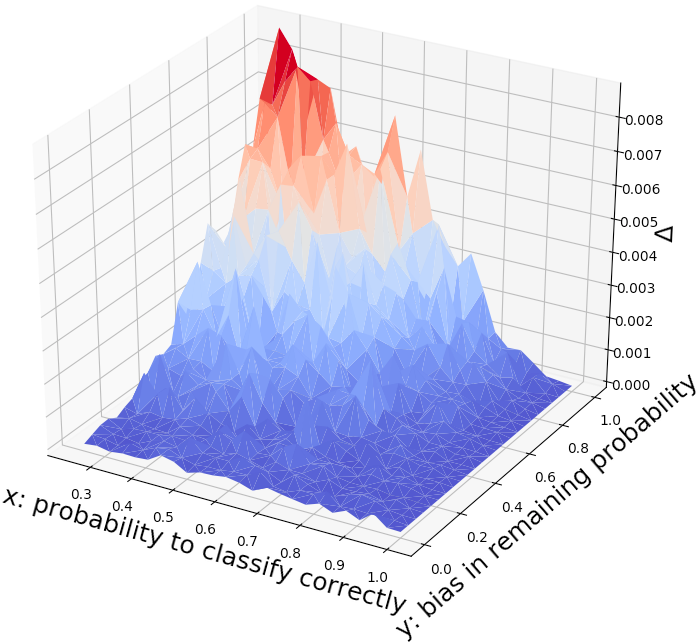}
    \caption{n=4}
    \label{fig:sf3}
    \end{subfigure}%
    \begin{subfigure}{.5\linewidth}
      \centering
      \includegraphics[width=\textwidth]{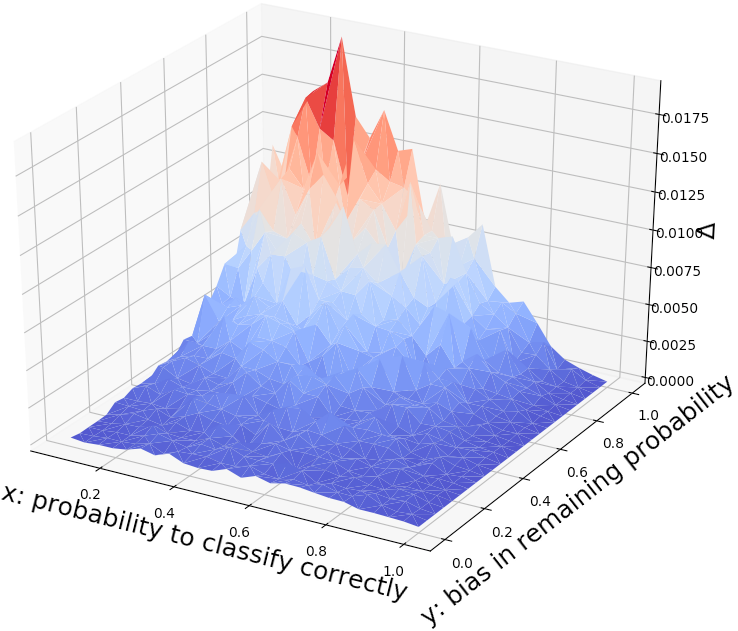}
    \caption{n=13}
    \label{fig:sf4}
    \end{subfigure} 
    \caption{ $\Delta$ for random classification tasks on 2000 data points with different classifier performance ($x$-axis) and classification probability for non-gold labels ($y$-axis).}
     \label{fig:ranclasbalanced}
\end{figure}

\paragraph{Balanced data sets}
Figures \ref{fig:sf3} and \ref{fig:sf4} show $\Delta$ for random classification tasks with varying classifier performance on balanced label distributions.
The $x$-axis represents the probability that data points are classified correctly (ranging from $\frac{1}{n}$ to 1).
With the $y$-axis we control the classification probability for remaining classes, ranging from $\frac{1-x}{n-1} [y=0]$ to $j\frac{1-x}{\sum_j j-i}[y=1]$, where $i$ is the true label.
 We find differences of up to 0.8 percentage points for n=4 and 1.7 percentage points for n=13.

\section{Discussion and conclusion}

Two formulas for calculating `macro F1' are found in the literature.
When precision and recall do not differ much within classes, the difference between evaluating a classifier with one or the other metric is negligible.
However, we can easily see cases where the outcomes diverge and are vastly different.
More specifically, we find that one metric ($\mathbb{F}_1$) is overly `benevolent' towards heavily biased classifiers and can yield misleadingly high evaluation scores.
This is likely to happen when the data set is imbalanced.
Moreover, the two macro F1 scores may not only diverge in their absolute score but also lead to different classifier rankings.
Since macro F1 is often used with the intention to assign equal weight to frequent and infrequent classes, we recommend evaluating classifiers with $\mathcal{F}_1$ (the arithmetic mean over individual F1 scores), which is significantly more robust towards the error type distribution. At the very least, researchers should indicate which formula they are using.

\bibliographystyle{plain}
\bibliography{references}

\begin{thebibliography}{1}

\bibitem{lipton2014optimal}
Zachary~C Lipton, Charles Elkan, and Balakrishnan Naryanaswamy.
\newblock Optimal thresholding of classifiers to maximize f1 measure.
\newblock In {\em Joint European Conference on Machine Learning and Knowledge
  Discovery in Databases}, pages 225--239. Springer, 2014.

\bibitem{opitz-frank-2019-argument}
Juri Opitz and Anette Frank.
\newblock An argument-marker model for syntax-agnostic proto-role labeling.
\newblock In {\em Proceedings of the Eighth Joint Conference on Lexical and
  Computational Semantics (*{SEM} 2019)}, pages 224--234, Minneapolis,
  Minnesota, June 2019. Association for Computational Linguistics.

\bibitem{rosenthal2015semeval}
Sara Rosenthal, Preslav Nakov, Svetlana Kiritchenko, Saif Mohammad, Alan
  Ritter, and Veselin Stoyanov.
\newblock Semeval-2015 task 10: Sentiment analysis in twitter.
\newblock In {\em Proceedings of the 9th international workshop on semantic
  evaluation (SemEval 2015)}, pages 451--463, 2015.

\bibitem{rudinger-etal-2018-neural}
Rachel Rudinger, Adam Teichert, Ryan Culkin, Sheng Zhang, and Benjamin
  Van~Durme.
\newblock Neural-davidsonian semantic proto-role labeling.
\newblock In {\em Proceedings of the 2018 Conference on Empirical Methods in
  Natural Language Processing}, pages 944--955, Brussels, Belgium,
  October-November 2018. Association for Computational Linguistics.

\bibitem{santos2011comparative}
A~Santos, A~Canuto, and Antonino~Feitosa Neto.
\newblock A comparative analysis of classification methods to multi-label tasks
  in different application domains.
\newblock {\em Int. J. Comput. Inform. Syst. Indust. Manag. Appl}, 3:218--227,
  2011.

\bibitem{sokolova2009systematic}
Marina Sokolova and Guy Lapalme.
\newblock A systematic analysis of performance measures for classification
  tasks.
\newblock {\em Information processing \& management}, 45(4):427--437, 2009.

\bibitem{wu2017unified}
Xi-Zhu Wu and Zhi-Hua Zhou.
\newblock A unified view of multi-label performance measures.
\newblock In {\em Proceedings of the 34th International Conference on Machine
  Learning-Volume 70}, pages 3780--3788. JMLR. org, 2017.

\end{thebibliography}

\newpage
\begin{appendices}
\section{Proof Lemma}
\label{proof:lemma}
Let $\sigma=n\sum_x(P_x +R_x)$. All summations exclude classes where $P_i+R_i=0$. 
\begin{equation}
\begin{split}
	\mathbb{F}_1 - \mathcal{F}_1
	&= \frac{2*\frac{1}{n}(\sum_xP_x)*\frac{1}{n}(\sum_y R_y)}{\frac{1}{n}\sum_x(P_x+R_x)} 
	- \frac{1}{\sum_x(P_x + R_x)} \mathcal{F}_1 \sum_y(P_y + R_y)\\
	&= \frac{2}{n\sum_x P_x+R_x}\sum_ {x,y} P_xR_y - \frac{1}{\sum_x(P_x + R_x)} \frac{2}{n} \sum_x \frac{P_xR_x}{P_x + R_x} \sum_y(P_y + R_y)\\
    &= \frac{2}{\sigma}(\sum_ {x,y} P_xR_y - \sum_ {x,y} \frac{P_xR_x (P_y + R_y)}{P_x + R_x})\\
&= \frac{2}{\sigma} \sum_{x\neq y} \frac{P_xR_y(P_x + R_x) - P_xR_x(P_y + R_y)}{P_x + R_x} \text{ [zero for $x=y$]}\\
&= \frac{2}{\sigma} \sum_{x<y} (\frac{P_xR_y (P_x + R_x) - P_x R_x (P_y + R_y)}{P_x + R_x} + \frac{P_yR_x (P_y + R_y) - P_y R_y (P_x + R_x)}{P_y + R_y})\\
&=\frac{2}{\sigma} \sum_{x<y} \frac{P_x^2 R_y^2 - 2P_x R_y P_yR_x +P_y^2 R_x^2 }{(P_x +R_x)(P_y +R_y)} \\
&=\frac{1}{\sigma} \sum_{x, y} \frac{(P_x R_y - P_y R_x)^2 }{(P_x +R_x)(P_y +R_y)} \qed
\end{split}
\end{equation}

\section{Proof Theorem 2.}
\begin{enumerate}
\item (i) $\Rightarrow$ (ii):
	$\text{Assume }\nexists i\in 1...n: P_i \neq R_i \xRightarrow[]{\text{Lemma (4)}}  \mathbb{F}_1-\mathcal{F}_1 = 0\ \lightning$
\item (ii) $\Rightarrow$ (iii): W.l.o.g. $P_i > R_i (\Leftrightarrow\sum_ym_{iy} < \sum_xm_{xi})$. Assume $\nexists j\in 1...n:P_j<R_j\ (\Leftrightarrow \nexists j\in 1...n:\sum_ym_{jy} > \sum_xm_{xj}) \Rightarrow \sum_i\sum_y m_{iy} < \sum_i\sum_x m_{xi}\ \lightning
$
\item (iii) $\Rightarrow$ (i):
	$(P_i R_j - P_j R_i)^2 > 0 \xRightarrow[]{\text{Lemma (4)}} \mathbb{F}_1-\mathcal{F}_1 > 0 \qed$
\end{enumerate}

\ \\
\section{Proof Theorem 3.}\label{sec:proof23} Preliminaries:
Consider the extended set of Precision-Recall-Configurations $Q = [0, 1]^{2\times n}$ and the discrete boundary set $Q^* = \{(0,1), (1,0)\}^{2\times n}$. Note that not all $q\in Q$ are realisable by a confusion matrix. It suffices to show that
\begin{enumerate}
	\item $\forall q\in Q: \exists q^* \in Q^*: \Delta^{q^*} \geq \Delta^q$
	\item max$_{q\in Q^*}(\Delta^{q})=\begin{cases}
    0.5,n\text{ is even}\\
    0.5 - \frac{1}{2n^2},\text{else}
	\end{cases}$
	\item max$_{q\in Q^*}(\Delta^q)$ can be be approximated by a sequence of suitable confusion matrices.
\end{enumerate}

\ \\
Note that for any fixed $i$,  $\Delta$ can be written as follows: \begin{displaymath}
\begin{split}
	\Delta
&= \frac{1}{n(\underbrace{\sum_{\substack{x\\x\neq i}}(P_x +R_x)}_{\alpha_i}+P_i+R_i)}
(\underbrace{\sum_{\substack{x\neq y\\x,y \neq i}} \frac{(P_xR_y - P_yR_x)^2 }{(P_x+R_x)(P_y+R_y)}}_{\beta_i}
+\underbrace{2\sum_{\substack{x\\x\neq i}} \frac{(P_iR_x-P_xR_i)^2 }{(P_i+R_i)(P_x +R_x)}}_{\gamma_i})\\
&= \frac{1}{n(\alpha_i+P_i+R_i)}(\beta_i+\gamma_i)
\end{split}
\end{displaymath}

\ \\\ \\
Proof 1.:\\
We construct $q^*\in Q^*$ in two steps:\\
(i) Iterate over all classes. If both $P_i$ and $R_i$ are non-zero, set $P_i$ or $R_i$ to 0 depending on the configuration of the remaining classes.\\
(ii) Set all non-zero variables to 1.\\
(iii) Iterate over all classes. If $P_i=R_i=0$, set $P_i$ to 1.

\ \\
(i)
Let $q=(P_1, R_1, ..., P_n, R_n) \in Q$.\\
Iteratively $\forall i$ where $P_i, R_i\neq(0,0)$: Determine the condition under which $P_i, R_i$ can be swapped in order to increase $\Delta$. Let $\tilde{q_i}=(P_1, R_1, ..., R_i, P_i, ..., P_n, R_n)$:
\begin{equation}
\begin{split}
	\Delta^{\tilde{q_i}} - \Delta^q
&= \frac{1}{n(\alpha_i + R_i + P_i)}
(\beta_i +2\sum_{\substack{x\\x\neq i}} \frac{(R_iR_x-P_xP_i)^2}{(R_i+P_i)(P_x +R_x)})\\
&- \frac{1}{n(\alpha_i + P_i + R_i)}
(\beta_i +2\sum_{\substack{x\\x\neq i}} \frac{(P_iR_x-P_xR_i)^2}{(P_i+R_i)(P_x +R_x)})\\
&= \frac{2}{n(\alpha_i + P_i + R_i)}
\sum_{\substack{x\\x\neq i}} \frac{(R_iR_x)^2+(P_xP_i)^2 - (P_iR_x)^2 - (P_xR_i)^2}{(P_i+R_i)(P_x +R_x)}\\
&= \frac{2}{n(\alpha_i + P_i + R_i)} \frac{R_i^2-P_i^2}{P_i+R_i}
\underbrace{\sum_{\substack{x\\x\neq i}} \frac{R_x^2-P_x^2}{P_x +R_x}}_{\delta_i}
\end{split}
\end{equation}
$\Delta^{\tilde{q_i}} - \Delta^q>0$ iff $P_i, R_i$ are skewed in the same direction as $\delta_i$.
In this case, swap $P_i, R_i$.
Let $q=(P_1, R_1, ..., P_n, R_n)$ henceforth denote the new configuration after a possible swap.
Now, $(R_i-P_i)\delta_i\leq 0$.
Proceed with a case distinction to set $P_i$ or $R_i$ to zero. (For $R_i-P_i=\delta_i=0$, both cases are possible.) \\

1. Case: $R_i\leq P_i$ and $\delta_i\geq0$. Set $R_i\rightarrow 0$. Let $q^i= (P_1, R_1, ..., P_i, 0, ..., P_n, R_n)$:
\begin{equation}
\begin{split}
\Delta^{q^i} - \Delta^q &=\frac{1}{n(\alpha_i+P_i)}(\beta_i
	+ 2\sum_{\substack{x\\x\neq i}} \frac{(P_iR_x)^2}{P_i(P_x+R_x)}) \\
&- \frac{1}{n(\alpha_i+P_i+R_i)}(\beta_i+2\sum_{\substack{x\\x\neq i}} \frac{(P_iR_x-P_xR_i)^2 }{(P_i+R_i)(P_x+R_x)}) \\
&\geq \frac{2}{n(\alpha_i+P_i+R_i)} \sum_{\substack{x\\x\neq i}} \frac{(P_iR_x)^2(P_i+R_i) - P_i((P_iR_x)^2 - 2P_iR_xP_xR_i + (P_xR_i)^2) }{P_i(P_i+R_i)(P_x +R_x)}\\
&\geq \frac{2}{n(\alpha_i+P_i+R_i)} \sum_{\substack{x\\x\neq i}} \frac{(P_iR_x)^2R_i - P_i(P_xR_i)^2}{P_i(P_i+R_i)(P_x +R_x)}\\
&\geq \frac{2}{n(\alpha_i+P_i+R_i)} \sum_{\substack{x\\x\neq i}} \frac{P_iR_i^2R_x^2 - P_iR_i^2P_x^2}{P_i(P_i+R_i)(P_x +R_x)}\\
&\geq \frac{2}{n(\alpha_i+P_i+R_i)} \frac{R_i^2}{P_i+R_i} \sum_{\substack{x\\x\neq i}} \frac{R_x^2-P_x^2}{P_x+R_x} \geq 0
\end{split}
\end{equation}

2. Case: $R_i\geq P_i$ and $\delta_i\leq0$. Set $P_i\rightarrow 0$. Let $q^i= (P_1, R_1, ..., 0, R_i, ..., P_n, R_n)$:
\begin{equation}
	\begin{split}
	\Delta^{q^i} - \Delta^q &=\frac{1}{n(\alpha_i+R_i)}(\beta_i
	+ 2\sum_{\substack{x\\x\neq i}} \frac{(-P_xR_i)^2}{R_i(P_x+R_x)})
	- \frac{1}{n(\alpha_i+P_i + R_i)}(\beta_i+\gamma_i)\\
	&\geq... 
	\geq \frac{2}{n(\alpha_i+P_i+R_i)} \frac{P_i^2}{P_i+R_i}
	\underbrace{\sum_{\substack{x\\x\neq i}} \frac{P_x^2-R_x^2}{P_x +R_x}}_{-\delta_i}
	\geq 0
	\end{split}
	\end{equation}

\ \\
(ii) Let $q=(P_1, R_1, ..., P_n, R_n)$ where $\forall i: P_i=0 \lor R_i=0$. Then
\begin{displaymath}
	\Delta = \frac{1}{n\sum_i(P_i+R_i)}\sum_{i,j}(P_iR_j+P_jR_i) = \frac{2}{n\sum_i(P_i+R_i)}\sum_{i,j}P_iR_j
\end{displaymath}
$\Delta$ can be increased by setting all non-zero variables to 1, since for any set of positive real-valued variables $x_1,...,x_{n_x},y_1,...,y_{n_y}$:
\begin{displaymath}
	\frac{\partial}{\partial x_i}\underbrace{\frac{\sum_{i,j}x_iy_j}{x_1+...+x_{n_x}+y_1+...+y_{n_y}}}_{\sigma}= \frac{(\sum_jy_j)\sigma - \sum_{i,j}x_iy_i}{\sigma^2} > 0.
\end{displaymath}
Analogously $\frac{\partial}{\partial y_i}$.

\ \\
(iii)
Let $q=(P_1, R_1, ..., P_n, R_n)$ where $\forall i: (P_i,R_i) \in \{(0,0), (0,1), (1,0)\}$.\\
Iteratively $\forall i$ where $P_i=R_i=0$:
Let $r=|\{i:(P_i,R_i)=(0,1)\}|$, $s=|\{i:(P_i,R_i)=(1,0)\}|$.
Let $q_i=(P_1, R_1, ..., 1, 0, ..., P_n, R_n)$ 
\begin{displaymath}
\begin{split} \Delta^{q_i}-\Delta^q
	&= \frac{2}{n(r+s+1)} r(s+1) - \frac{2}{n(r+s)}rs\\
	&= \frac{2}{n}\frac{(rs+r)(r+s) - rs(r+s+1)}{(r+s+1)(r+s)}\\
	&= \frac{2}{n} \frac{r^2}{(r+s+1)(r+s)} \geq 0
\end{split}
\end{displaymath}
\qed

\ \\
Proof 2.:\\
Let $q\in Q^*$ and $r,s$ as defined above. Note that $\mathcal{F}_1=0$ and $r+s=n$.
$$\Rightarrow \Delta = \mathbb{F}_1 = \frac{2 \frac{r}{n} \frac{s}{n}}{\frac{r}{n}+\frac{s}{n}} = \frac{2}{n^2}rs
$$
$\Delta$
is maximised for $r,s=\frac{n}{2}$ ($n$ is even) or $r,s=\frac{n-1}{2},\frac{n+1}{2}$ (else).
$$\Rightarrow \Delta^{max} =
\begin{cases}
    \frac{1}{2}, n\text{ is even}\\
    \frac{1}{2}(1 - \frac{1}{n^2}), \text{ else}
\end{cases}
$$
\qed

\ \\
Proof 3.\\
For any fixed $n\geq2$, let $(m^n)_{z\in\mathbb{N}_0}$ be a sequence of confusion matrices with $$ m^n_z =
\begin{pmatrix}
	1 & &\text{\huge{0}}\\
	z & 1\\
	\text{\huge{0}} & &\ddots
\end{pmatrix}
\text{ (n is even), }\indent
\begin{pmatrix}
	1 & & & & &\text{\huge{0}}\\
	z & 1\\
	& & \ddots\\
	& & & 1\\
	&  & & z & 1 & z\\
	\text{\huge{0}} & & & & & 1
\end{pmatrix}
\text{ (else)}$$
Then $q^{m^n_z} = (1, \frac{1}{1+z}, \frac{1}{1+z}, 1, ...)$ or 
$(..., 1, \frac{1}{1+z}, \frac{1}{1+2z}, 1, 1, \frac{1}{1+z})$
with $\lim\limits_{z\rightarrow\infty}{q^{m^n_z}} = (1,0, 0, 1, ...)$
and
$\lim\limits_{z\rightarrow\infty}{\Delta^{m^n_z}} = \Delta^{max}.$
\qed\\
\ \\
$\qed$

\section{Implementation example}

\begin{table}[!htb]
    \begin{minipage}{.5\linewidth}
      \centering
         \begin{tabular}{c|c|c}
        &a     & b\\
        \hline
       a& 100 & 10,000  \\
       b& 0 & 100 \\
    \end{tabular}
    \caption{$\mathbb{F}_1$ = 0.505 \textcolor{red}{\bm{$\gg$}} $\mathcal{F}_1$ = 0.0196. Note that individual F1 score=0.0196 for both classes.}
    \label{tab:subtabx1}
    \end{minipage}%
    \begin{minipage}{.5\linewidth}
      \centering
        \begin{tabular}{c|c|c}
    &a     & b\\
        \hline
      a &  100 & 5,000  \\
      b & 5,000 & 100 
    \end{tabular}
    \caption{$\mathbb{F}_1$ = 0.0196 \textcolor{calpolypomonagreen}{\bm{$\equiv$}} $\mathcal{F}_1$ = 0.0196}
    \label{tab:subtabx2}
    \end{minipage} 
     \caption{One macro F1 metric ($\mathbb{F}_1$) is very sensitive towards the error type distribution, while the other is not ($\mathcal{F}_1$). }
     \label{tab:exx1}
\end{table}

Compile the script in Appendix \ref{app:B}:

\begin{verbatim}
    $ ghc mf1.hs
\end{verbatim}

Input: Number of classes and a confusion matrix. For example, to calculate the scores for two classes and a confusion matrix $[[100,10000],[0,100]]$:
\begin{verbatim}
    $ ./mf1 2 100 10000 0 100
\end{verbatim}
This prints:
\begin{verbatim}
    ("macroF1 benevolent",0.504950495049505)
    ("macroF1 non-benevolent",1.96078431372549e-2)
    ("delta",0.48534265191225007)
    ("delta calculated",0.48534265191225007)
\end{verbatim}

The result for the confusion matrix with `balanced' error type distribution $[[100,5000],[5000,100]]$:
\begin{verbatim}
    ("macroF1 benevolent",1.96078431372549e-2)
    ("macroF1 non-benevolent",1.96078431372549e-2)
    ("delta",0.0)
    ("delta calculated",0.0)
\end{verbatim}

\section{ Example code}
\enlargethispage{100pt}

\label{app:B}

\begin{Verbatim}[fontsize=\tiny]
import System.Environment
import Data.List
import Control.Applicative

type CellIdx = (Int, Int)

crossProduct :: [a] -> [(a,a)]
crossProduct xs = (,) <$> xs <*> xs

valueAt :: CellIdx -> [[a]] -> a
valueAt (i,j) xss = xss !! i !! j

pairs :: [a] -> [(a, a)]
pairs xs = [(x,y) | (x:ys) <- tails xs, y <- ys]

diag :: [[a]] -> [a]
diag xss = zipWith (!!) xss [0..]

rowSum :: (Num a) => Int -> [[a]] -> a
rowSum i xss = sum $ xss !! i

colSum :: (Num a) => Int -> [[a]] -> a
colSum i xss = sum $ ( transpose xss ) !! i

rec :: (Fractional a) => Int -> [[a]] -> a
rec i xss = (/) ( valueAt (i,i) xss ) $ colSum i xss 

harMean :: (Fractional a) => a -> a -> a
harMean x y = (*2) $ (/) ( x * y ) ( x + y )

f1 :: (Fractional a) => Int -> [[a]] -> a
f1 i xss = harMean p r 
    where 
        p = prec i xss
        r = rec i xss

macroF1 :: (Fractional a) => [[a]] -> a
macroF1 xss = harMean ( avgPrec xss ) ( avgRec xss )

macroF1' :: (Fractional a) => [[a]] -> a
macroF1' xss = (/)
                ( sum [f1 i xss | i <- [0..(length xss)-1]] )
                ( fromIntegral . length $ xss )

prec :: (Fractional a) => Int -> [[a]] -> a
prec i xss = (/) ( valueAt (i,i) xss ) $ rowSum i xss 

avgPrec :: (Fractional a) =>  [[a]] -> a
avgPrec xss = (/) 
                ( sum [prec i xss | i <- [0..(length xss)-1]] ) 
                ( fromIntegral . length $ xss )

diffForTuple :: (Fractional a) => CellIdx -> [[a]] -> a
diffForTuple (i,j) xss =  ( 2 * (P_x * R_y - P_y*R_x)^2 ) / normalizer 
    where 
        normalizer = ( fromIntegral $ length xss ) 
                     * ( P_x + R_x ) * (P_y + R_y) 
                     * ( sum [
                              (prec k xss) + (rec k xss) 
                              | k <- [0..(length xss)-1]
                             ]  
                       )
        P_x = prec i xss
        R_x = rec i xss
        R_y = rec j xss
        P_y = prec j xss

deltaF1_F1' :: (Fractional a) => [[a]] -> a
deltaF1_F1' xss = sum $ [ diffForTuple pair xss 
                          | pair <- pairs [i | i <-[0..(length xss) -1 ]] ]

 
avgRec :: (Fractional a) =>  [[a]] -> a
avgRec xss = (/) 
            ( sum [rec i xss | i <- [0..(length xss)-1]] ) 
            ( fromIntegral . length $ xss )

cm :: Int -> [a] -> [[a]]
cm i [] = []
cm i xs = [take i xs] ++ ( cm i $ drop i xs )

ri :: String -> Int
ri i = read i


main = do
     args <- getArgs
     let is = map ri args
     let f1 =  macroF1 . cm (head is) . map fromIntegral $ drop 1 is
     let f2 =  macroF1' . cm (head is) . map fromIntegral $ drop 1 is
     print $ ("macroF1 benevolent", f1)
     print $ ("macroF1 non-benevolent", f2)
     print $ ("delta",f1 - f2)
     print $ ("delta calculated", deltaF1_F1' . cm (head is) . map fromIntegral $ drop 1 is)
\end{Verbatim}


\end{appendices}

\end{document}